\documentclass{article}
\usepackage{spconf,amsmath,graphicx}
\usepackage{multirow,tabularx,booktabs,array}
\usepackage{color,soul}
\usepackage{xcolor}

\title{Contrastive Siamese Network for Semi-supervised Speech Recognition}
\name{Soheil Khorram${^*}$, Jaeyoung Kim${^*}$, Anshuman Tripathi, Han Lu, Qian Zhang, Hasim Sak\thanks{${^*}$ These authors contributed equally to this work}}
\address{\{soheilkhorram, jaeykim, anshumant, luha, zhaqian, hasim\}@google.com\\Google Inc., USA}

\newcommand{\per}[1]{#1}
\newcommand{\bper}[1]{\textbf{#1}}
\begin{document}
\ninept
\maketitle
\begin{abstract}
This paper introduces contrastive siamese (\emph{c-siam}) network, an architecture for leveraging unlabeled acoustic data in speech recognition. \emph{c-siam} is the first network that extracts high-level linguistic information from speech by matching outputs of two identical transformer encoders. It contains augmented and target branches which are trained by: (1) masking inputs and matching outputs with a contrastive loss, (2) incorporating a stop gradient operation on the target branch, (3) using an extra learnable transformation on the augmented branch, (4) introducing new temporal augment functions to prevent the shortcut learning problem. We use the Libri-light 60k unsupervised data and the LibriSpeech 100hrs/960hrs supervised data to compare c-siam and other best-performing systems. Our experiments show that c-siam provides $20\%$ relative word error rate improvement over wav2vec baselines. A c-siam network with 450M parameters achieves competitive results compared to the state-of-the-art networks with 600M parameters.

%Experimental results show that c-siam provides $20\%$ relative word error rate improvement over wav2vec baselines. Our experiments show that c-siam achieves competitive results compared to current state-of-the-art systems, with a smaller network, when using the Libri-light 60k unsupervised data and the LibriSpeech 960 hrs supervised data.
\end{abstract}
\vspace{-3pt}
\section{Introduction}
\label{sec:intro}
\vspace{-3pt}

Collecting large transcribed datasets is expensive and time consuming. Also, because of privacy concerns of the users, we always prefer to minimize transcribing datasets while keeping the quality of the systems unchanged. A common method to this end is to leverage unlabeled data using self/semi-supervised techniques. This work is an attempt to improve the performance of the existing self/semi-supervised speech recognition techniques. 
% We introduce contrastive siamese (c-siam) network, the fist architecture that extracts high-level linguistic information from speech by matching outputs of audio encoders.

Most approaches in learning effective representations without human supervision fall into one of three categories that predict (1) input-level, (2) intermediate-level and (3) output-level representations. It is easier to train the first category as there is no degenerate solutions for them. For example, in autoregressive predictive coding (APC) \cite{chung2019unsupervised} the goal is to generate future frames by conditioning on past frames using a unidirectional network; or similarly DeCoAR \cite{ling2020deep,ling2020decoar}, TERA \cite{liu2021tera} and MOCKINGJAY \cite{liu2020mockingjay} are systems that mask inputs and generate the mask regions with a bidirectional network. All these methods benefit from L1 reconstruction loss which is stable and easy to optimize. 
% Generating input features enables them to capture different levels of speech information and therefore these methods can be used for various speech processing tasks, e.g., speech recognition and speaker verification \cite{chung2019unsupervised}. 
However, there is a major problem: In order to generate inputs, the network needs to learn details of the inputs that are not necessary for supervised tasks. As a result, it is not optimal to incorporate these techniques alongside supervised losses in semi-supervised frameworks.

The second category is based on predicting intermediate-level representations. Some methods in this category include CPC \cite{oord2018representation}, wav2vec \cite{schneider2019wav2vec} and vq-wav2vec \cite{baevski2019vq} which are similar to unidirectional APC \cite{chung2019unsupervised}, but instead of generating future inputs they predict future intermediate representations. Wav2vec 2.0 \cite{baevski2020wav2vec} and w2v-BERT \cite{chung2021w2v} extend these unidirectional structures to bidirectional ones. They mask the intermediate representations and predict the masked regions. 
%It is not practical to train these systems with L1 or other generative losses, because they will collapse to a trivial solution (e.g., constant embeddings for all inputs); 
These techniques incorporate contrastive \cite{gutmann2010noise} or clustering \cite{caron2018deep} losses which are more consistent with supervised tasks, but there is still room for improvement; it is more effective to predict outputs of audio encoders \cite{chen2020simple}.

Speech SimCLR \cite{jiang2020speech,chen2020simple} uses both input and output level prediction losses. An augmentation module transforms inputs into two correlated views, then a transformer extracts output-level representations for both views. Speech SimCLR minimizes two losses: a contrastive loss that matches output representations, and a reconstruction loss that matches inputs and outputs. The reconstruction loss is employed to prevent the \emph{shortcut learning problem} \cite{geirhos2020shortcut}: SimCLR can easily minimize the contrastive loss by ignoring the inputs and by using positional embeddings of the transformers. However, this reconstruction loss prevents SimCLR to be consistent with supervised tasks. In this paper, we propose to use temporal augmentation to deal with the shortcut learning problem. We introduce two temporal augmentation methods that are consistent with speech recognition and they can effectively reduce the shortcut learning problem. 

To evaluate the efficacy of the c-siam network, we developed semi-supervised speech recognition systems using the Libri-light 60k unsupervised data and the LibriSpeech 100hrs/960hrs supervised data. We compared c-siam with current state-of-the-art methods and the results show that c-siam provides $20\%$ relative word error rate (WER) improvement when compared with the wav2vec-based self/semi-supervised systems. 
% The results also show that c-siam, with a 450M parameter network, can match the current best systems with 600M parameters.

\begin{figure*}[t]

\begin{minipage}[b]{1.0\linewidth}
  \centering
  \centerline{\includegraphics[width=17.8cm]{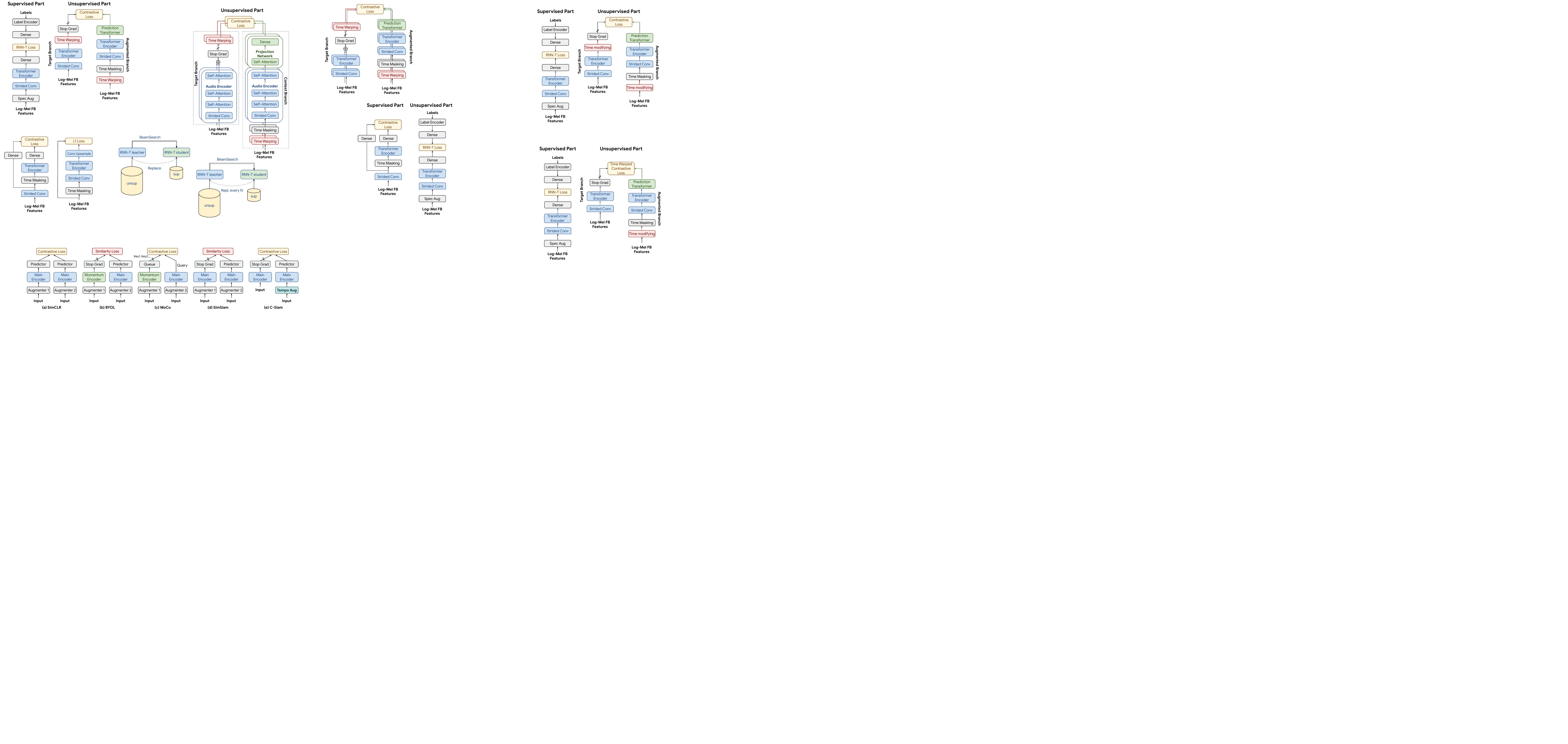}}
%  \vspace{2.0cm}
\end{minipage}

\footnotesize{
\hspace{23pt}
(a) SimCLR~\cite{chen2020simple} \hspace{45pt}
(b) BYOL~\cite{grill2020bootstrap} \hspace{50pt}
(c) MoCo~\cite{he2020momentum} \hspace{48pt}
(d) SimSiam~\cite{chen2021exploring} \hspace{50pt}
(e) C-Siam}
\vspace{-5pt}

\caption{Architecture of self/semi-supervised networks that are based on the idea of matching encoders' outputs. 
% (a) SimCLR: simple framework for contrastive learning of representations \cite{chen2020simple}, (b) BYOL: bootstrap your own latent \cite{grill2020bootstrap}, (c) MoCo: momentum contrast \cite{he2020momentum}, (c) SimSiam: simple siamese network \cite{chen2021exploring}, and (d) the proposed c-siam network. ``TempoAug'' is our proposed temporal augmentation module.
}
\label{fig:comparison}
\vspace{-15pt}
\end{figure*}

\vspace{-3pt}
\section{Related Work}
\label{sec:related work}
\vspace{-3pt}

Our work is most related to SimCLR~\cite{chen2020simple}, BYOL~\cite{grill2020bootstrap}, MoCo~\cite{he2020momentum} and SimSiam~\cite{chen2021exploring} methods. Figure \ref{fig:comparison} shows the architecture of these methods compared to c-siam. They all follow the core idea of matching high-level representations generated from two branches. A trivial solution of matching high-level representations is all outputs ``\emph{collapsing}'' to a constant vector~\cite{chen2021exploring}. In this section, we discuss how these methods prevent the collapsing problem. 

\emph{SimCLR} learns representations by maximizing agreement between two different augmentations of the same data sample. It benefits from a contrastive loss to prevent the collapsing problem. 
% By separating positive and negative samples through the contrastive loss, SimCLR branches do not converge to a constant. 
SimCLR experiments showed that a learnable nonlinear mapping on top of the encoders can significantly improve the quality of the representations~\cite{chen2020simple}. \emph{BYOL} relies on similarity losses, but it does not collapse as it uses a momentum encoder. It contains online and target branches.
% which work on two augmentations of the inputs. 
The online branch is trained to predict the target outputs and the target branch is an exponential average of the online branch. BYOL also incorporates a learnable predictor on the online branch~\cite{chen2020simple}. \emph{MoCo} leverages both the contrastive loss and the momentum encoder. It considers contrastive learning as a dictionary look-up such that the main encoder extracts a query representation and the momentum encoder extracts a queue of keys. MoCo uses the contrastive loss to match queries with their corresponding keys~\cite{he2020momentum}. \emph{SimSiam} uses the same encoders in both branches and it matches the encoders' outputs using the cosine similarity function.
%instead of the contrastive loss. 
SimSiam deals with the collapsing problem by applying a stop gradient operation and a learnable projection network~\cite{chen2021exploring}.

Although these methods are effective for learning from unlabeled images, they are not suitable for transformer-based speech recognition.
% as they cannot handle sequences of embeddings generated by transformers. 
The main issue is the \emph{``shortcut learning problem''} \cite{geirhos2020shortcut}: when we process sequences of inputs, 
% positional embeddings are enough to match the outputs and therefore 
transformers tend to minimize the training loss by just using the positional information and by ignoring the inputs. To overcome this problem, we propose to use temporal augmentation in c-siam. We modify the temporal characteristics of the inputs before processing them by the encoders. We also modify our contrastive loss to align the representations before defining positive and negative pairs.
% Therefore, in c-siam, the encoders take inputs with different time characteristics and they cannot cheat by just considering the positional information.

\vspace{-3pt}
\section{Preliminary Experiment}
\label{sec:inconsistency}
\vspace{-3pt}

We first conduct an experiment to figure out if wav2vec-style training~\cite{baevski2020wav2vec} is consistent with supervised speech recognition. The experiment consists of two steps: (1) we train an audio encoder using the wav2vec 2.0 scheme on the Libri-light 60k data; (2) we extract intermediate representations of the encoder and train a simple classifier on each representation to recognize frame-level phonemes. We expect to see constant improvement in phoneme recognition accuracy when we get closer to the output of the encoder.

Our audio encoder starts with two layers of strided convolutions, each of which downsamples its input by the factor of two. The encoder follows by a 20-layer transformer-xl with relative attention mechanism~\cite{dai2019transformer}. We train this encoder with the wav2vec 2.0 scheme explained in~\cite{zhang2020pushing}.
% Wav2vec 2.0 scheme masks the input of the audio encoder and tries to predict the mask regions using the contrastive loss introduced in~\cite{baevski2020wav2vec}. Masking is done by simply setting the embeddings to zero. We use $100$ negative samples in our contrastive loss, the length of the mask regions is $10$ and the probability of masking each frame is $0.065$, which are all consistent with~\cite{chung2021w2v}.
In order to perform phoneme recognition, we apply two dense and a softmax layers after each transformer layer. We train them with the cross-entropy (CE) loss that maximizes the likelihood of predicting frame-level phonemes. 
% This CE loss just trains the dense layers and it does not affect parameters of the main encoder. 
We follow~\cite{oord2018representation} for preparing our frame-level phoneme recognition dataset.
% We divide LibriSpeech 100 hrs dataset into $80\%$ train and $20\%$ test sets to train and evaluate our phoneme recognition layers. Frame-level Phonemes are generated with the same procedure explained in~\cite{oord2018representation}.

Figure 2 (red, square points) shows phoneme recognition results achieved by wav2vec training. The accuracy increases from $65.9\%$ at layer $1$ to $91.2\%$ at layer $17$. In the last 3 layers, the accuracy drops to $70.2\%$. This performance reduction is a result of wav2vec trying to match the inputs of the audio encoder and it is not easy to predict phonemes from the inputs. To address this problem, we propose to match higher-level representations using siamese networks.

We also repeat the above experiment with our c-siam network (see Section~\ref{sec:csiam} for the details of c-siam). We just use unsupervised part of the c-siam network in this experiment to be consistent with the wav2vec experiment. Figure 1 (blue, circular points) shows c-siam's phoneme recognition results. It does not show any performance drop at the end of the audio encoder. The constant improvement in phoneme accuracy confirms the consistency between the c-siam structure and the phoneme recognition task, and it motivates us to use c-siam alongside supervised speech recognition losses.

\begin{figure}[t]

\begin{minipage}[b]{1.0\linewidth}
  \centering
  \centerline{\includegraphics[width=7cm]{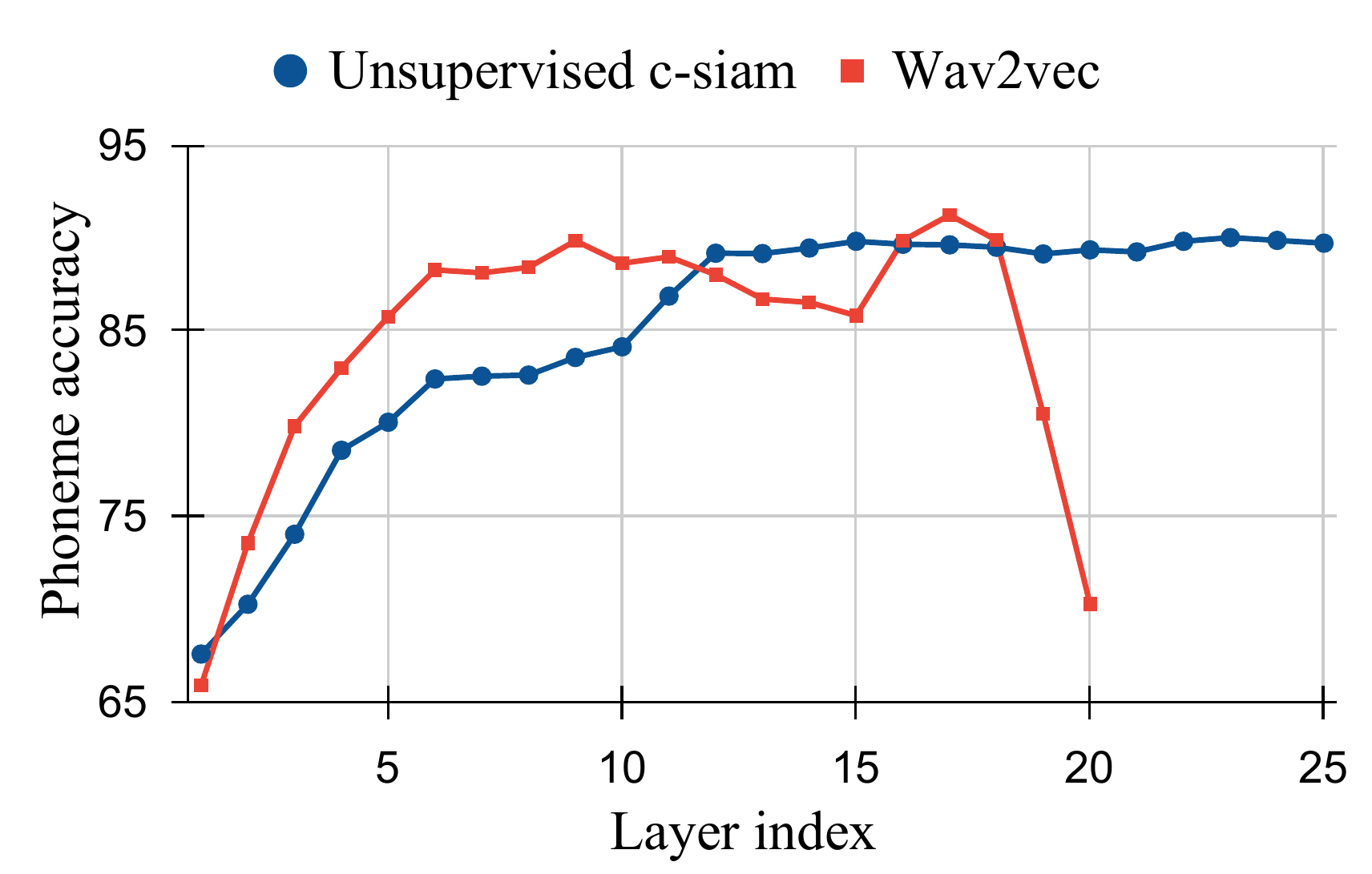}}
  \vspace{-8pt}
\end{minipage}
\caption{Frame-level phoneme recognition accuracy using two dense layers for both wav2vec 2.0 (red, square points) and unsupervised c-siam (blue, circular points). 
% We calculate accuracy for all layers of the the audio encoder. The accuracy drops significantly for the last layers of the wav2vec.
\vspace{-15pt}
}
\label{fig:res}
\end{figure}

% Phoneme recognition experiment for wav2vec

\vspace{-3pt}
\section{Contrastive siamese network}
\label{sec:csiam}
\vspace{-3pt}

\begin{figure}[t]

\begin{minipage}[b]{1.0\linewidth}
  \centering
  \centerline{\includegraphics[width=7.5cm]{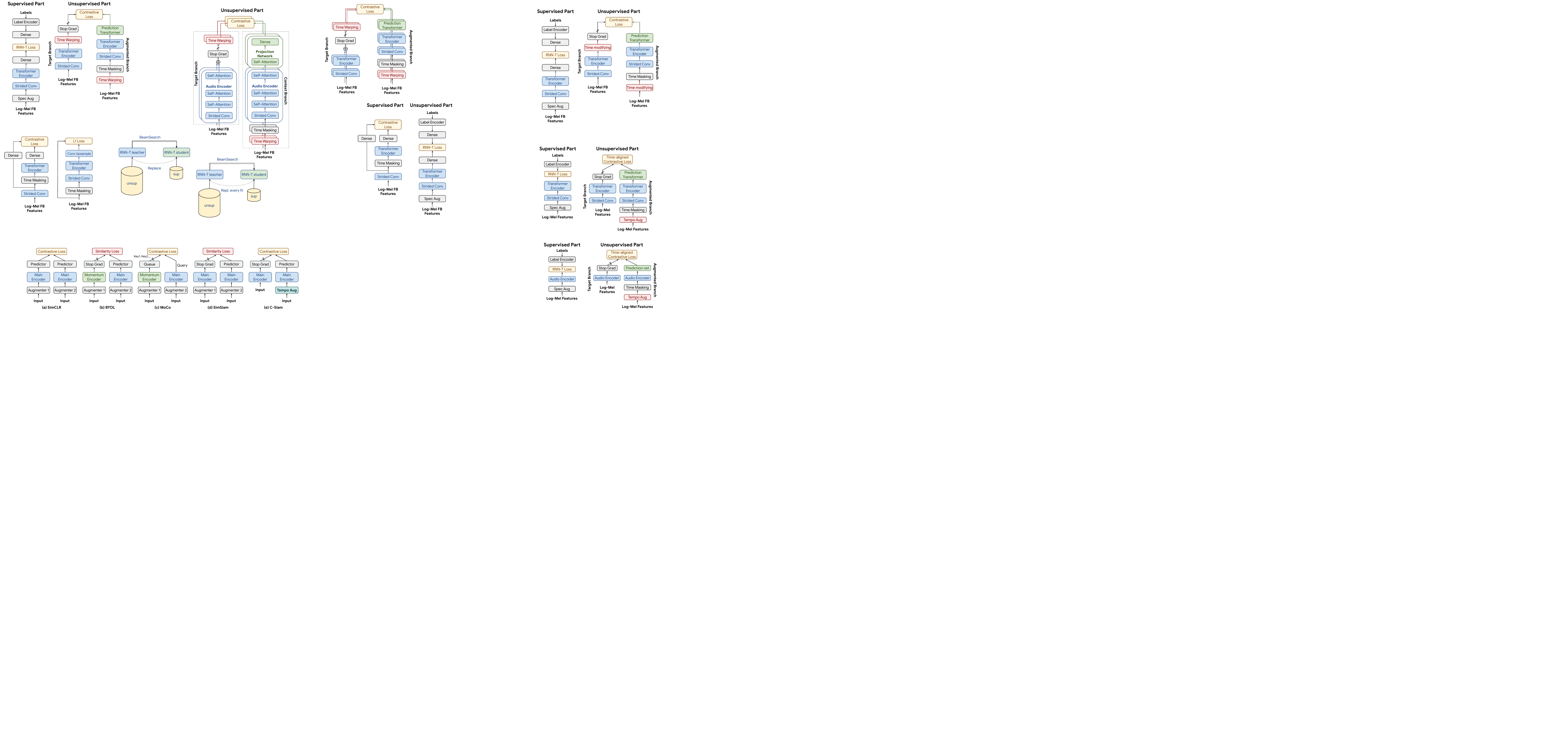}}
  \vspace{-5pt}
\end{minipage}
\caption{Supervised and unsupervised parts of the c-siam network.}
\vspace{-15pt}
\label{fig:arch}
\end{figure}

An illustration of the c-siam network is shown in Figure~\ref{fig:arch}. The network contains supervised and unsupervised parts. 
Both parts share the same audio encoder and they are trained together using the same Adam optimizer and the same learning rate function. 
% We explain different components of each part in the following subsections.

\vspace{-5pt}
\subsection{Supervised Network}
\label{sec:supervised}
\vspace{-3pt}

Supervised network is consistent with the RNN-T based transformer transducer structure introduced in~\cite{zhang2020transformer}. In this structure, likelihood of labels given input features are factorized through three modules: an audio encoder, a label encoder and a logit function. 
\emph{Audio encoder} is a stack of strided convolution layers followed by transformers. Two strided convolution layers downsample log-mel features by the factor of four. Then, multiple layers of transformer-xl~\cite{dai2019transformer} extract our acoustic embeddings. \emph{Label encoder} is a streaming transformer-xl that does not attend to the future labels. 
% Each layer is a multi-head attention that 
%first normalizes inputs with a LayerNorm, then 
% projects inputs to Query, Key, and Value vectors. Query and Key are combined with positional embeddings using a relative attention mechanism~\cite{dai2019transformer} to generate attention weights. These weights are used to calculate weighted average of the Value vectors, which generates the outputs of the heads. Finally, these outputs are concatenated and are passed to a dense layer. \anshumant{No need to go into this much detail given that we have already cited the paper}
% The details of our transformer layers are explained in \cite{zhang2020transformer}.
% \emph{Label encoder} is a streaming transformer-xl. It is similar to the audio encoder, but with two main differences: (1) instead of strided convolutions it uses an embedding layer that generates a dense vector for each label; (2) it does not attend to the future labels by setting the attention weights of the future labels to zero. 
% \anshumant {Maybe remove this too, we can just say in a single line audio and label encoder are transformer-xl layers}
\emph{Logit function} takes both acoustic and label embeddings as inputs and generates logit embeddings using this module:
\vspace{-3pt}
\begin{equation}
    \vspace{-2pt}
    r = \mathrm{Linear}(\mathrm{Tanh}(\mathrm{Linear}(a) + \mathrm{Linear}(l))),
    \label{eq:logit}
    % \vspace{-2pt}
\end{equation}
where $a$, $l$ and $r$ are acoustic, label and logit embeddings respectively. 
% $\mathrm{Linear}$ and $\mathrm{Tanh}$ are conventional neural network layers. 
% We pass the logit vectors to a softmax layer to calculate the probability of the labels (graphemes and RNN-T's blank symbol).
We pass the logits to a softmax and calculate label probabilities which are used in RNN-T's forward/backward algorithm~\cite{graves2012sequence}.

% In order to reduce the overtraining problem, both audio and label encoders benefit from different regularization techniques: dropping out intermediate activations, adding variational noise to the weights, contributing L2 loss of the weights in training. Furthermore, input features are augmented with the specaugment (SpecAug) method introduced in~\cite{park2019specaugment}.

\vspace{-5pt}
\subsection{Unsupervised Network}
\label{sec:unsupervised}
\vspace{-3pt}

% We train unsupervised network with unlabeled data samples. 
Unsupervised network contains two branches: augmented and target. \emph{Augmented branch} takes log-mel features and applies temporal augmentation (TempoAug), time masking, audio encoder and prediction network to predict the outputs of the target branch. The \emph{target branch} extracts clean outputs by passing log-mel features to the audio encoder. It is designed to just generate the clean outputs and we do not train parameters through that. To do so, we apply a stop gradient operation at the end of the target branch.

\emph{Stop gradient and prediction network} -- These components are introduced in SimSiam architecture~\cite{chen2021exploring} to improve convergence properties of siamese networks. By using these components, target network generates expected outputs based on the knowledge learned up to the current state of training, and the augmented branch tries to match these expected outputs~\cite{chen2021exploring}. We use 5 layers of transformer-xl as the prediction network in our experiments.

\emph{Time-aligned contrastive loss} -- In order to match target and augmented outputs, we mask the features of the augmented branch and we minimize a contrastive loss over the masked regions. 
% Both masking and calculating the contrastive loss are similar to wav2vec 2.0~\cite{baevski2020wav2vec}. 
Masking is done by simply setting continuous regions of features to zero, and contrastive loss is a softmax-based negative log-likelihood function defined over cosine similarities. Assume $a_t$ is an output vector of the augmented branch,
% generated from a masked feature vector
$q_{t'}$ is a positive target vector that has to match $a_t$, and $Q$ is a set of negative target vectors randomly selected from the masked regions of the same utterance. Contrastive loss at time $t$ is:
\\
\vspace{-4pt}
\begin{equation}
    \mathcal{L}_t =
    - \log
    \frac{
        \exp (\mathrm{sim}(a_t, q_{t'}) / \tau)
    }
    {
        \sum_{q \in Q \cup \{q_{t'}\}}{\exp (\mathrm{sim}(a_t, q) / \tau)}
    },
    \label{eq:contrastive_loss}
\end{equation} 
\vspace{-4pt}
\\
where $\mathrm{sim}$ is the cosine similarity and $\tau$ is the temperature parameter.
% and the summation is taken over the union of $Q$ and $\{q_{t'}\}$. 
In c-siam, positive pairs may come from different time indexes ($t$ and $t'$ in Eq.~\ref{eq:contrastive_loss}). It is because of the temporal augment (TempoAug) module used at the beginning of the augmented branch. 

\emph{TempoAug} -- modifies temporal characteristics of the inputs by shifting log-mel features in time. The goal is to prevent \emph{``shortcut learning problem''} \cite{geirhos2020shortcut} caused by transformers' positional embeddings. 
% Transformers benefit from positional embeddings to capture temporal dynamics of their inputs. 
If we do not modify time characteristics of the branches, network can ignore its input and minimize the contrastive loss just based on the positional embeddings. We introduce two TempoAug techniques in this work: uniform and non-uniform.
% We explain them in the rest of this section.

\emph{Uniform TempoAug} -- uniformly compresses or stretches time-domain audio signals with a randomly drawn tempo ratio. It uses a waveform similarity based overlap and add (WSOLA)~\cite{verhelst1993overlap} method to linearly change tempo of speech (i.e., changing $t$ to ${\alpha}t$) without modifying its pitch contours. In our experiments, the ratio $\alpha$ is randomly drawn from a uniform distribution for each utterance. Since each utterance takes different tempo ratio, our audio encoder cannot easily model it and therefore positional counting can be avoided. 

% when it is applied to one of branches in time-aligned contrastive loss.
% For an input audio signal, $x(t)$, our uniform TempoAug changes it to $x(\alpha t)$, where $\alpha$ is randomly drawn for each utterance. Since each utterance takes different tempo ratio, positional counting can be avoided when it is applied to one of branches in time-aligned contrastive loss.

\emph{Non-uniform TempoAug} -- non-uniformly compresses or stretches log-mel trajectories such that it modifies temporal characteristics, but it does not negatively affect speech recognition. We do this by applying a time-warping function to feature trajectories. Assume $x(t)$ denotes speech features at time $t$, we transform it to $x(w(t))$, where $w(t)$ denotes our time-warping function. We also apply the same time-warping function to the outputs of the target branch in our time-aligned contrastive loss. This warping function must satisfy three constraints to preserve the nature of log-mel trajectories: (1) Monotonicity -- $w(t)$ must be monotonically increasing, otherwise it will not preserve the order of the input samples. (2) Smoothness -- sudden changes in the warping function distort the overall shape of the feature trajectories. (3)  Boundary conditions -- the warping function must start from time $0$, $w(0) = 0$, and it must end to the last frame, $w(T-1) = T-1$, where $T$ is the number of input frames. Boundary condition ensures that the warping function does not eliminate any part of the inputs. We propose the following time-warping function that satisfies all these constraints:
\vspace{-7pt}
\begin{equation}
    w(t) = t + \sum_{r=1}^{R} a_r \mathrm{sin}(\frac{\pi r t}{T - 1}),\ \ t \in \{ 0, 1, ..., T-1\}.
    \label{eq:warping_function}
    \vspace{-5pt}
\end{equation} 
In this equation, $R$ is the order of warping function and $a_r$ specifies the amplitude of the $r$-th $\mathrm{sin}$ component. These parameters can control smoothness and monotonicity of the warping function. In our experiments, we use five $\mathrm{sin}$ components ($R=5$) and we randomly select the values of $a_r$ for each utterance from the normal distribution with mean $0$ and std $0.2$. We empirically found that these parameters lead to our best performing systems.

After generating a time-warping function, we must apply it to the input features, i.e., we must calculate $x(w(t))$. We use a linear interpolation technique to do this:
% However, calculating $x(w)$ is not straightforward as $w$ is not an integer index and it may be any real value in the range $[0, T-1]$. To deal with this issue, we linearly interpolate the values of $x(w)$:
\vspace{-5pt}
\begin{equation}
    \vspace{-2pt}
    x(w) \approx (w - \lfloor w \rfloor)x(\lceil w \rceil) + (\lceil w \rceil - w)x(\lfloor w \rfloor),
    \label{eq:linear_interpolation}
    % \vspace{-1pt}
\end{equation} 
where $\lfloor w \rfloor$ and  $\lceil w \rceil$ are floor and ceil values of $w$. Our experiments showed that applying time-warping function introduced in Eq.~\ref{eq:warping_function} using the simple linear interpolation expressed by Eq.~\ref{eq:linear_interpolation} can significantly reduce the shortcut learning problem.

\vspace{-2pt}
\section{Experiments}
\label{sec:experiments}
\vspace{-3pt}

% We conduct several experiments to compare c-siam with other self/semi-supervised techniques.
% improve RNN-T based transformer transducer speech recognizers using unlabeled data.

\emph{Data} --
We evaluate the c-siam network using the LibriSpeech dataset~\cite{panayotov2015librispeech}. It is a read speech corpus collected from LibriVox audio books. We use two standard sets, 100 hrs and 960 hrs sets, of LibriSpeech as our supervised datasets. Our unsupervised dataset is the Libri-light corpus which is also derived from the LibriVox audio books. Libri-light contains 60k hours of 16kHz audio. Our language model (LM) training dataset is a combination of LibriSpeech text transcripts with 10M word tokens and an additional text-only dataset with  800M word tokens. We extract 80-dimensional 10ms log-mel filter bank coefficients as our acoustic features~\cite{zhang2020pushing}. Also, we use SpecAugment~\cite{park2019specaugment} for the supervised part of the network. SpecAugment applies two frequency masks with the size of 27 and 10 time masks with the maximum ratio of 0.05.

\emph{Models} -- 
We train two types of models, a small model with 100M parameters and a large model with 450M parameters. The small model contains 2 conv layers with the kernel size of 3 and 512 channels, 20 layers of transformers, 8 attention heads, 512 dimensional outer embeddings, 2048 dimensional inner embeddings, 48 dimensional Value vectors and 16 dimensional Query/Key vectors. The large model contains 24 transformer layers, 16 heads, 1024 dimensional outer embeddings, 4096 dimensional inner embeddings and other parameters are the same as the small model. We train our c-siam models on 16x16 TPUs with a per-core batch size of 2 (experiments on 100hrs) and 4 (experiments on 960hrs). This results in a global batch size of 1024 and 2048. Both supervised and unsupervised utterances share the same batch size in our c-siam network.

\emph{Training parameters} -- 
We train both models with the Adam optimizer~\cite{kingma2014adam}. Learning rate is ramped up linearly to 2e-3 during first 10k steps, and then decays exponentially to 2.5e-6 at 200k steps. We also apply gradient scaling to limit the norm of the gradient vectors to 60. To reduce the over-training problem, we set dropout factor, variational noise power and L2 loss weight to 0.3, 0.02 and 1.5e-4 for our small model and we set them to 0.1, 0.03 and 3e-5 for our large model. We apply the variational noise after 4k iterations. To do masking we sample initial time steps of the masks randomly with probability 0.016 and we mask the subsequent 28 steps.

% It also provides a larger unlabeled dataset from LibriVox, the Libri-light set, which is roughly 60k hours of audio. In our experiments, we use the Libri-light set as our unsupervised set and a 960-hour subset of the transcribed LbriSpeech as our supervised set.
% Google Voice Search corpus consists of more than 30k hours of speech. Its test set contains 14k Voice Search utterances with the maximum length of 5.5 seconds per utterance. The training and test sets are all anonymized and hand-transcribed. 
% In order to train our networks, we first extract 80 dimensional log-mel features, we then apply two strided convolutional layers that subsample the features by the factor of four. This process results in a sequence of 40ms embeddings that form the inputs of our audio encoders.

\begin{table}[t]
  \centering
  \setlength{\tabcolsep}{3.5pt}
  \begin{tabularx}{0.46\textwidth}{l || c | c c | c c}
    \toprule
    \multicolumn{2}{c|}{} & \multicolumn{2}{c|}{Clean} & \multicolumn{2}{c}{Other} \\
    Method & Size (B) & Dev. & Test & Dev. & Test \\
    [2pt] \hline \hline \rule{0pt}{12pt}
    \hspace{-2.5pt}\footnotesize{\textbf{Baseline Models}} & & & & & \\
    random init. & .45 & \per{1.9} & \per{2.0} & \per{4.2} & \per{4.5}  \\
    wav2vec init. & .45 & \per{1.9} & \per{1.9} & \per{4.1} & \per{4.0}  \\
    wav2vec cotrain & .45 & \per{1.7} & \per{1.8} & \per{3.8} & \per{3.5} \\
    [2pt] \hline \rule{0pt}{12pt}
    \hspace{-2.5pt}\footnotesize{\textbf{Proposed Models}} & & & & & \\
    uniform c-siam & \multirow{2}*{.45}  & \per{1.6} & \per{1.7} & \per{2.9} & \bper{2.8} \\
    uniform c-siam + LM & & \per{1.5} & \per{1.6} & \per{2.7} & \bper{2.8} \\
    non-uniform c-siam & \multirow{2}*{.45} & \per{1.5} & \per{1.6} & \per{2.7} & \per{2.9} \\
    non-uniform c-siam + LM &  & \bper{1.4} & \per{1.6} & \bper{2.6} & \bper{2.8}  \\
    [2pt] \hline \rule{0pt}{12pt}
    \hspace{-2.5pt}\footnotesize{\textbf{SOTA Models}} & & & & & \\
    NST-conformer \cite{zhang2020pushing} & .1  & \per{1.6} & \per{1.7} & \per{3.3} & \per{3.5} \\
    w2v-CTC \cite{baevski2020wav2vec} & .3  & \per{2.1} & \per{2.2} & \per{4.5} & \per{4.5} \\
    w2v-conformer \cite{zhang2020pushing} & .6  & \per{1.7} & \per{1.7} & \per{3.5} & \per{3.5} \\
    w2v-BERT \cite{chung2021w2v} & \multirow{2}*{.6}  & \per{1.5} & \bper{1.5} & \per{2.9} & \per{2.9} \\
    w2v-BERT + LM \cite{chung2021w2v} &  & \bper{1.4} & \bper{1.5} & \per{2.8} & \bper{2.8} \\
    \bottomrule
  \end{tabularx}
  \caption{WER results on LibriSpeech 960 hrs data. The size of the networks are reported in billion parameters. Also, rows with ``+ LM'' report the results obtained after applying a language model.
%   Baseline section refers to the existing models trained by us using our RNN-T and transformer-xl structure. State-of-the-art (SOTA) section reports the best results from the literature.
}
  \label{tab:wer_result_960}
  \vspace{-12pt}
\end{table}

\vspace{-5pt}
\subsection{Experiment Results}
Table \ref{tab:wer_result_960} and \ref{tab:wer_result_100} report the WER results when LibriSpeech 960 hrs and 100 hrs are used as the supervised data. The WER numbers are calculated on the standard LibriSpeech development and test sets. Both tables are divided into three sections: baseline, proposed and state-of-the-art (SOTA). 
We trained our transformer-xl network using the existing self/semi-supervised methods and we summarize the results in the baseline section. In this section, \emph{``Random init.''} refers to the base system that does not leverage any unlabeled data, \emph{``wav2vec init.''} is considered for the pretraining/finetuning experiment with wav2vec 2.0, \emph{``Wav2vec cotrain''} is a semi-supervised model in which both RNN-T and wav2vec losses are trained together in a similar way that we train our c-siam network. In the proposed section, we report the WER numbers obtained for the c-siam network with both uniform and non-uniform TempoAug approaches. We also report our LM fusion results. We follow transformer-based shallow LM fusion explained in~\cite{zhang2020transformer} to obtain these numbers. The last section, SOTA section, contains WER of the best-performing systems that we could find from other papers. There are many differences between these systems and our c-siam system, for example ``NST-conformer'', ``w2v-conformer'' and ``w2v-BERT'' use the conformer encoders \cite{gulati2020conformer} instead of the transformer-xl and ``w2v-CTC'' is trained with CTC loss instead of the RNN-T loss. However, all these systems are similar in one aspect that they are trained with the same supervised and unsupervised datasets.

The results reported in the tables show that (1) c-siam significantly improves all the baseline models on the ``other'' set. The best c-siam network provides $15\%$ (on 100 hrs) and $20\%$ (on 960 hrs) relative WER improvement compared to the best baseline system (i.e., ``wav2vec cotrain''). (2) The improvement achieved by c-siam on the ``clean'' set ($11\%$ for 960 hrs and $3\%$ for 100 hrs) is less than the ``other'' set. This is also True when we compare ``wav2vec cotrain'' and ``wav2vec init.''. It is due to the nature of our unsupervised data that is more similar to the ``other'' set; therefore, all cotrain systems are more effective for the ``other'' set. (3) c-siam with 450M parameters achieves competitive results compared to the best performing models in the literature with 600M parameters. This is an important achievement for practical applications in which there is a limitation in using large models. (4) Uniform and non-uniform TempoAug perform similarly in 960 hrs experiments, but non-uniform TempoAug is better in 100 hrs. It shows more augmentation provided by the non-uniform TempoAug is helpful for smaller datasets. LibriSpeech is divided into two separate evaluation sets: ``Clean'' and 'Other'. The ``Other'' set is noisier and therefore it contains more challenging utterances. C-Siam provides more improvement in the ``Other'' set, because C-Siam's target embeddings are close to the output labels and they are less impacted by the input noise.

There is an important limitation with the c-siam architecture. In each training step, c-siam needs to run the forward pass of the audio encoder three times, which leads to a slow training loop and also high memory usage. Training our large network with c-siam takes 5 days, while with ``wav2vec cotrain'' it takes 3.5 days. Due to this problem, it is challenging to train c-siam for very large networks.

\begin{table}[t]
  \centering
  \setlength{\tabcolsep}{4.5pt}
  \begin{tabularx}{0.46\textwidth}{l || c | c c | c c}
    \toprule
    \multicolumn{2}{c|}{} & \multicolumn{2}{c|}{Clean} & \multicolumn{2}{c}{Other} \\
    Method & Size (B) & Dev. & Test & Dev. & Test \\
    [2pt] \hline \hline \rule{0pt}{12pt}
    \hspace{-2.5pt}\footnotesize{\textbf{Baseline Models}} & & & & & \\
    random init. & .1 & \per{5.3} & \per{5.5} & \per{16.} & \per{15.8}  \\
    wav2vec init. & .1 & \per{3.1} & \per{3.0} & \per{7.3} & \per{6.8}  \\
    wav2vec cotrain & .1 & -- & \per{3.2} & -- & \per{6.2} \\
    [2pt] \hline \rule{0pt}{12pt}
    \hspace{-2.5pt}\footnotesize{\textbf{Proposed Models}} & & & & & \\
    uniform c-siam (S) & .1  & \per{3.0} & \per{2.9} & \per{5.3} & \per{5.3} \\
    non-uniform c-siam (S) & .1 & \per{2.9} & \per{3.0} & \per{5.2} & \per{5.3}  \\
    % [2pt] \hline \rule{0pt}{12pt}
    % \footnotesize{\textbf{Our Large Models}} & & & & & \\
    uniform c-siam (L) & .45  & \per{3.0} & \per{2.9} & \per{4.6} & \per{4.8} \\
    non-uniform c-siam (L) & .45  & \per{2.7} & \per{2.6} & \bper{4.3} & \bper{4.6} \\
    [2pt] \hline \rule{0pt}{12pt}
    \hspace{-2.5pt}\footnotesize{\textbf{SOTA Models}} & & & & & \\
    w2v-CTC \cite{baevski2020wav2vec} & .3  & \per{3.3} & \per{3.1} & \per{6.5} & \per{6.3} \\
    w2v-conformer \cite{zhang2020pushing} & .6  & \per{2.5} & \per{2.6} & \per{4.7} & \per{4.9} \\
    w2v-BERT \cite{chung2021w2v} & .6  & \bper{2.4} & \bper{2.5} & \per{4.4} & \bper{4.6} \\
    \bottomrule
  \end{tabularx}
  \caption{WER results for LibriSpeech 100 hrs data. ``(S)'' and ``(L)'' are used to refer to small and large c-siam networks.
  % Uniform and non-uniform specify the type of the TempoAug technique used in training c-siam network. Also, (S) and (L) refer to the size of the c-siam network.
  }
  \label{tab:wer_result_100}
  \vspace{-15pt}
\end{table}

\vspace{-5pt}
\section{Conclusion}
\label{sec:conclusion}

\vspace{-5pt}
We introduce contrastive siamese (c-siam) network, a new architecture for training semi-supervised speech recognition systems. c-siam simultaneously trains a supervised RNN-T model and an unsupervised siamese network. The siamese network contains target and augmented branches. It extracts clean and augmented representations from target and augmented branches; it then modifies the augmented representations to be correlated with the clean ones using a contrastive loss. We train c-siam on 60k hours Libri-Light unlabeled data and LibriSpeech labeled data. We show that c-siam either outperforms or matches state-of-the-art systems. For future work, we plan to explore different techniques to train large networks (with 600M parameters or more) using c-siam.

\bibliographystyle{IEEEbib}
\bibliography{refs}

\end{document}